\documentclass[journel]{IEEEtran}
\IEEEoverridecommandlockouts
\usepackage{times}  
\usepackage{helvet}  
\usepackage{courier}  
\usepackage[hyphens]{url}  
\usepackage{graphicx} 
\usepackage{caption} 
\DeclareCaptionStyle{ruled}{labelfont=normalfont,labelsep=colon,strut=off} 
\usepackage{multirow}
\usepackage{adjustbox}
\usepackage{amsmath}
\usepackage{amssymb}
\usepackage{subfigure}
\usepackage{booktabs}
\usepackage{comment}
\usepackage[normalem]{ulem}
\usepackage[ruled,linesnumbered]{algorithm2e}
\usepackage[dvipsnames]{xcolor}

\usepackage{amssymb}

\newif\ifmodify

\ifmodify
\newcommand{\cross}[1]{\textcolor{red}{\sout{#1}}}

\newcommand{\fei}[1]{\textcolor{orange}{#1}}
\newcommand{\mh}[1]{\textcolor{purple}{#1}}
\else
\newcommand{\cross}[1]{}

\newcommand{\fei}[1]{#1}
\newcommand{\mh}[1]{#1}
\fi

\newif\ifmaximal

\title{Compact Multi-level Sparse Neural Networks with Input Independent Dynamic Rerouting}
\author{
    \IEEEauthorblockN{Minghai Qin\textsuperscript{1*}, Tianyun Zhang\textsuperscript{2*}, Fei Sun\textsuperscript{3*}, Yen-Kuang Chen\textsuperscript{3}, Makan Fardad\textsuperscript{4}, \\Yanzhi Wang\textsuperscript{5}, Yuan Xie\textsuperscript{3}
    \thanks{\textsuperscript{*}Equal contribution. }
    \thanks{\textsuperscript{1}Work done while working at Alibaba DAMO Academy. }
    \thanks{\textsuperscript{2}Work done while interning at Alibaba DAMO Academy.}\\}
    \IEEEauthorblockA{\textsuperscript{1}Western Digital Research \\
    \textsuperscript{2}Cleveland State University \\
    \textsuperscript{3}Alibaba DAMO Academy \\
    \textsuperscript{4}Syracuse University \\
    \textsuperscript{5}Northeastern University
    }
}

\usepackage{bibentry}

\begin{document}

\maketitle

\begin{abstract}
  Deep neural networks (DNNs) have shown to provide superb performance in many real life applications, but their large computation cost and storage requirement have prevented them from being deployed to many edge and internet-of-things (IoT) devices. Sparse deep neural networks, whose majority weight parameters are zeros, can substantially reduce the computation complexity and memory consumption of the models. 
  In real-use scenarios, devices may suffer from large fluctuations of the available computation and memory resources under different environment, and the quality of service (QoS) is difficult to maintain due to the long tail inferences with large latency.
  Facing the real-life challenges, we propose to train a sparse model that supports multiple sparse levels. That is, a \textbf{hierarchical structure} of weights are satisfied such that the locations and the values of the non-zero parameters of the more-sparse sub-model are a subset of the less-sparse sub-model.
  In this way, one can dynamically select the appropriate sparsity level during inference, while the storage cost is capped by the least sparse sub-model. 
  We have verified our methodologies on a variety of DNN models and tasks, including the ResNet-50, PointNet++, GNMT, and graph attention networks.
  \textcolor{black}{We obtain sparse sub-models with an average of 13.38\% weights and 14.97\% FLOPs, while the accuracies are as good as their dense counterparts. More-sparse sub-models with 5.38\% weights and 4.47\% of FLOPs, which are subsets of the less-sparse ones,  can be obtained with only 3.25\% relative accuracy loss.}
  In addition, our proposed hierarchical model structure supports the mechanism to inference the first part of the  model with less sparsity,
  and dynamically reroute to the more-sparse level if the real-time latency constraint is estimated to be violated. 
  \textcolor{black}{Preliminary analysis shows that we can improve the QoS by one or two nines depending on the task and the computation-memory resources of the inference engine.}
\end{abstract}

\section{Introduction} 
Deep neural networks (DNNs) have been widely used in many areas and have significantly changed our lives such as object recognition, autonomous driving, and  language translation. The fast development of DNNs is largely credited to the rapid growing of computation power, such as Graph Processing Units (GPUs) and Tensor Processing Units (TPUs), that enables the efficient search of the large and powerful DNN models.
On the other hand, there are a variety of environments that DNNs are deployed on the edge, where the computation capacity is limited. For example, mobile phones are increasingly integrating  more artificial intelligence features that require models to perform inference on a variety of tasks, such as picture beautification, image/video super resolution, 
and 3-D augmented reality. Internet-of-Things (IoTs) devices are becoming  more popular to be customized to understand the consumers' habits.
The state-of-the-art DNN models for these complicated tasks are usually extremely large, {\it i.e.}, containing huge amount of weights in the model, and thus not suitable to be directly deployed on mobile devices or embedded systems.

One of the widely-used approaches to cope with the complexity is to reduce the size of the models by sparsification: a technique to set the majority of the weights in the models to zero values (called {\em pruning}) through carefully designed algorithms, software, and hardware implementations~\cite{Han2016a,han2016eie}.
Since the pioneering work~\cite{Han2016a}, there are growing academic and industrial interests in this area.
The principle of DNN model sparsification lies in the ability to identify and prune the non-critical weights, while minimally reducing the accuracy loss  by adjusting  the values of the remaining weights. 


Real world applications in the wild need to handle dramatically different environments, which challenges the mobile inference engines~\cite{alibaba2020mnn}. For example, a mobile phone may allocate more computation capacity and power budget when it is fully charged and is at low temperature. But when its thermal limit is reached, the phone is forced to reduce the processor frequency to cool down the device, which is a process called {\it thermal throttling}. As the result, its computation capacity is significantly reduced. Thus, the inference latency distribution of DNN models may show two peaks, one  at its highest computation capacity, and one at the thermal throttled state~\cite{gaudette2018optimizing}.
In addition, an experiment has been performed in~\cite{wu2019machine} to measure the inference latency of a layer in a model on A11 iPhone mobile chipset. Even though the mean latency is 2.02ms with a standard deviation of 1.92ms, some inferences take 16ms or longer. These inferences with large latency, though small in percentage, may be easily visible to the customers. Thus, when a model is deployed to millions of devices, it is customary to optimize the 99\textsuperscript{th} or 99.9\textsuperscript{th} percentile latency to improve the quality of service (QoS). 
The traditional remedy to the long tail latency inferences is to skip frames in a series of inferences ({\it e.g.} in video analysis), allocating more time to the remaining inferences. This approach has an obvious drawback of the degraded quality visible to users.

As the result, one efficient sparse neural network model may still be insufficient to handle the changing environment with different optimization objectives. It is desirable to prepare two models: a relatively large model that is more accurate but cost more computation, and a relatively small model that is more efficient but less accurate. 
Admittedly, it is trivial to train two separate models at different sparsity levels and store them in the device. However, the storage and memory cost of the separate models are not satisfactory, which is more severe when more than two sparsity levels are desired.
It is much more appealing if multiple sparse models can be stored one on top of another, consuming no extra storage and memory compared to the least-sparse model. 

\begin{figure*}[t]
\centering
  \centering
  \includegraphics[width=\linewidth]{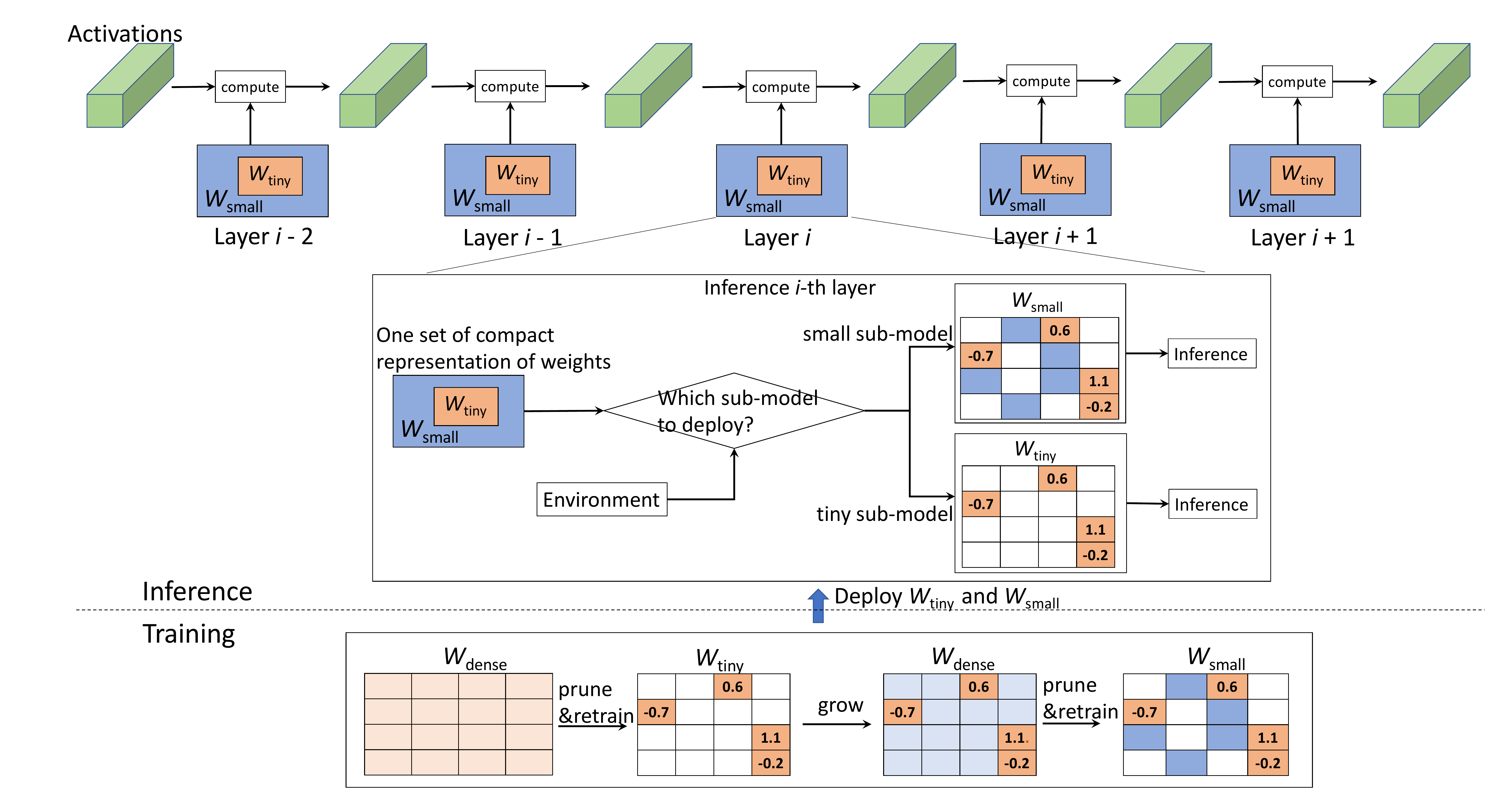}
  \caption{\small Inference and training methodology for the proposed compact multi-level sparse neural networks with weight hierarchy. During inference, the compute engine selects the desired sparsity level based on the environment. During training, a tiny-dense-small methodology is proposed to optimize tiny and small sub-models. Empty boxes denote zero-valued weights and colored boxes denote non-zero weights. Once the tiny sub-model is acquired, their weights are never changed in the rest of training process.}
  \label{fig:all}
\end{figure*}

In this work, we explore methodologies to address large variations of the inference latency on the devices in the wild {\it \textbf{within}} neural network models.
We propose a training and inference methodology that creates sparse neural network models with  multiple  sparsity  levels. The different sparsity levels follow a \textbf{\em hierarchical structure} such that the locations and values of the non-zero weights of the more-sparse sub-model is a subset of the less-sparse sub-model. 
We call the less-sparse sub-model as the {\textbf{\em small}} sub-model and the more-sparse sub-model as the \textbf{\em tiny} sub-model in this work.  
Figure~\ref{fig:all} shows an overview of our proposals. 
In the inference phase, the tiny sub-model (weights colored in orange) is a subset of the small one (weights colored in blue and orange) \textcolor{black}{and thus compact representations can be designed}. The appropriate sub-model may be selected based on the environment. For example, the small sub-model may be selected when the processor is at its full capacity, and the tiny sub-model is selected when the processor is at its thermal throttled state.
In the training phase,
we propose a ``tiny-dense-small'' training methodology that derives the small sub-model from the tiny sub-model. The resultant small sub-model performs as good as its dense counterpart, sometimes even better. 


With multiple levels of sparsity, the preferred level may be selected before each inference to reduce the the 99\textsuperscript {th} or 99.9\textsuperscript{th} percentile latency. 
For some critical applications with hard real-time constraint, making the selection before the inference may not be sufficient.
We  provide a mechanism to dynamically reroute the inference from the small to the tiny sub-models in the middle of the inference to further reduce the long tail latency. We also show that the accuracy after rerouting is better than the tiny sub-model. Depending on the rerouting locations, the accuracy improvement  may vary.

The contributions of the paper are summarized as follows.
\begin{itemize}
    \item  We propose a methodology to create sparse sub-models of a deep neural network at multiple sparsity levels, whose weights follow a logically hierarchical structure that the values and locations of the weights in the tiny sub-model are a subset of those in the small sub-model. This weight hierarchy brings 20\% to 30\% savings comparing with storing two separate sparse models.
    
    \item We present a physically hierarchical and compact sparse weight representation that has almost the same cost as storing the small sub-model, meanwhile we can efficiently extract from it all weights in the tiny sub-model. 
    
    \item We show that our methodology is universal and can be applied to different kinds of neural network models, such as 2-D image classification (ResNet-50), 3-D object recognition (PointNet++), machine translation (GNMT), and transductive learning (Graph attention networks).
    The results are reported as the geometric mean of the four models. We obtain small sub-models with 13.38\% weights and 14.97\% FLOPs with accuracy as good as their dense counterparts in~\cite{he2016deep,qi2017pointnet++,wu2016gnmt,velickovic2018gat}. Tiny sub-models, which are subsets of the small ones,  cost 5.38\% weights, 4.47\% of FLOPs, and result in only 3.25\% relative accuracy loss.
    

    \item We propose a dynamic inference rerouting method such that the model can partially inference the small sub-model in the beginning and switch to the tiny sub-model without throwing away the activation already computed. Experiments on ResNet-50 and GNMT have shown that the dynamic inference rerouting achieves higher accuracy than the tiny sub-model  at different rerouting positions.

    \item Our proposed multi-level sparsity solution has the potential to improve the QoS given a fixed latency bound. 
    \textcolor{black}{Based on the latency distribution from real devices, we have created a model to estimate the QoS for the small and tiny sub-models. With the same latency that the small sub-model delivers two nines of QoS (99\textsuperscript {th} percentile), the tiny sub-model is capable of delivering four nines of QoS (99.99\textsuperscript {th} percentile).  As far as we know, this is the first work looking into the problem on sparse models. }
    %
\end{itemize}

\section{Related Works}


Network pruning~\cite{hassibi1993second,LeCun1990BrainDamage} can reduce the  number of non-zero  weights  in  neural  network  models. Some works focus on {\em irregular} pruning where no constraints are imposed on the locations of non-zero weights.
~\cite{han2015learning} proposes a magnitude-based importance metric and iteratively prune weights with small magnitude.~\cite{LearningL02017iclr} uses $\ell_0$ regularization and~\cite{wen2016learning} manages to solve the non-convexity problem of  $\ell_0$ by $\ell_1$ approximation.~\cite{zhang2018systematic,ren2019admm} utilize an effective technique in optimization theory to improve the sparsity based on $\ell_0$ regularization. 
On the other hand, {\em structured} pruning imposes constraints on the locations of non-zero weights and thus reduces irregularity. 
For example, filter-wise pruning and shape-wise pruning~\cite{wen2016learning} can remove rows and columns in matrix-matrix multiplications, which translates to computation reduction in GPU platforms. Recent works explore some special dimensions and propose pattern pruning and kernel-wise pruning~\cite{ma2020pconv}. 
All previous works focus on a single sparse level and our work differs from them by creating multiple sparsity levels in a single model.
Our work uses ADMM~\cite{boyd2011distributed} \fei{algorithm} to \fei{irregularly} prune the dense model \cross{irregularly} to achieve the maximum accuracy.

Some other works propose to prune and grow the DNN back and forth. Transitioning from dense to sparse then back to dense has been shown to improve the accuracy of dense models~\cite{han2016dsd,Prakash2019trainConv}. In~\cite{Mallya2018packnet}, a dense network \cross{which is a union} \fei{composed} of multiple sparse networks is proposed to accomplish multiple tasks with different sparse subsets. 
While their goals are to train dense networks, our work differs in that the transition from tiny to dense is an intermediate step and our final goal is to obtain multiple sparse sub-models with weight hierarchy. 
~\cite{dai2019nest,dai2020GrowLSTM,Dai2019IncrementalLU} propose to grow part of connections based on the gradients of loss and prune it back to the desired sparsity. Similar efforts with different growing criterion are made in~\cite{Hassantabar2019SCANNSO,du2019cgap,Wen2019AutoGrowAL}.
While all the non-zero weights in the existing \cross{prune-and-grow or} grow-and-prune works are free to be updated, our work fundamentally differs in that the tiny sub-model is permanently frozen during all processes to guarantee the weight hierarchy.

\cite{onceForAll2020iclr} proposes an efficient progressive shrinking method to train a large super-network based on some architectures and deploy only a portion of it depending on the platform (mobile, cloud, or edge device). 
Other works~\cite{AdaptiveNN2017icml,SacrificingAF2018icann,MultiScaleDN2017iclr,shallowdeepnetworks2019icml,Teer16icpr} design to early exit based on some confidence score.
\textcolor{black}{While they focus on searching for dense networks, our work is orthogonal to them in the search of sparse sub-models. Another difference from~\cite{onceForAll2020iclr} is that our sparse weight hierarchy is a tight restriction while weights of two dense elastic kernels in~\cite{onceForAll2020iclr} are related by matrix transformations and are fine-tuned.}

\section{Methodology}

In this section, we present our training methodologies to acquire the small and tiny sub-models whose weights follow a hierarchical structure. 
We enforce the sparse weights in convolution, matrix-vector (MV) multiplication, and matrix-matrix (MM) multiplication operators to have multiple levels of sparsity. The weights in these operators represent the majority of the parameters in a model.
The biases in those operators, if exist, remain dense and are not constrained. 
The weights and biases in other operators, such as batch normalization are also not constrained. The parameters in those operators only contribute a tiny fraction of the model size and floating point operations (FLOPs).

Throughout the paper, we denote a DNN model (or a sub-model) by $M$, denote the set of non-zero weights of $M$ by $W(M)$, and denote the locations of the non-zero weights of $M$ by $L(M)$. The pair $(L(M), W(M))$ uniquely defines the weight structure of the DNN model $M$.

\subsection{A tiny-dense-small training process}

\begin{figure*}[tbp]
\centering
  \centering
  \includegraphics[width=\linewidth]{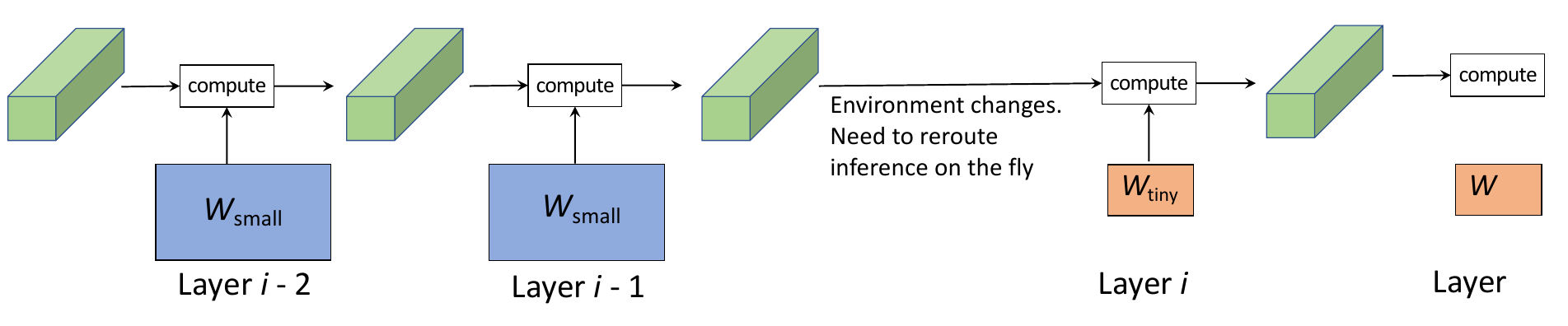}
  \caption{\small Dynamic rerouting of multi-level sparse neural networks from the small to the tiny sub-models.}
  \label{fig:dynamic} 
\end{figure*}

First, we pre-train a dense DNN model into a tiny one with good learning accuracy to start with. This is shown in the first step in the training phase of Figure~\ref{fig:all}. It is known that different layers may have different redundancies in maintaining the accuracy when they are sparsified~\cite{wen2016learning}. In this paper, we try to limit this degrees of freedom for a fair comparison, such that all layers in one sparse sub-model have the same sparsity level.
In order to obtain the tiny sub-model, we use the
alternating direction method of multipliers (ADMM)~\cite{boyd2011distributed} algorithm
that has been shown to perform well in restraining the number of non-zero weights in each layer. 

The second step is to grow the tiny sub-model back to a dense DNN model $M_d$ by fixing those weights $W(M_t)$ in the location $L(M_t)$. It is shown in the second step in the training phase of Figure~\ref{fig:all}. Note that the non-zero weights in the orange boxes are frozen in this step (and the third step as well).
All other weights are trainable and can be updated to maximally increase the accuracy of the dense DNN model. 
Note that the first two steps are similar to the dense-sparse-dense training method~\cite{han2016dsd} except for two differences. 
First, $W(M_t)$ in the location $L(M_t)$ of the dense model is permanently fixed to guarantee that the dense model is a superset of the tiny sub-model.
Second, the pruned weights in the tiny sub-model during re-training to the dense  model are sometimes required to be re-initialized as random numbers, as opposed to zero values in~\cite{han2016dsd}, since  some of our tiny sub-models are too sparse for many weights to be updated in the backward propagation. For example, if the entire kernel of a convolutional layer or the entire row of a weight matrix is pruned, re-training  by initializing  them as zeros will not update them effectively. 

The last step is to prune the dense DNN model back to a small sub-model. It is shown in the third step in the training phase in Figure~\ref{fig:all}. During this pruning, all weights $W(M_t)$ in the location $L(M_t)$ are fixed. It means that the values of weights $W(M_t)$ in the location $L(M_t)$ are never changed once we obtain the tiny sub-model. This is illustrated by the orange boxes of the last three weight matrices in the training phase in Figure~\ref{fig:all}. 
In this way, it is guaranteed that the final small sub-model $M_s$ satisfies $L(M_t) \subset L(M_{s})$ and $W(M_t)\subset W(M_s)$.
This pruning process again utilizes a modified version of the ADMM algorithm, where the projection of the dense tensor to the sparse tensor requires to keep $W(M_t)$ in $L(M_t)$ intact. 

After this step, we have \fei{the} tiny and small sub-models such that their weights follow a hierarchical structure, i.e., $L(M_t) \subset L(M_s)$ and $W(M_t) \subset W(M_s)$.

We use two-level sparsity to demonstrate our ``tiny-dense-small'' training methodology. 
\cross{In practice, two-level sparsity has demonstrated a good  trade-off between accuracy and DNN complexity.}
In order to obtain more than two levels, \cross{we can generalize } the training method \fei{can be generalized} to ``micro-dense-tiny-dense-small'' for triple-level sparsity, where the ``micro'' sub-model can be three to four times smaller than the tiny one. \fei{In practice, a} three to four times difference between \fei{the} adjacent sparsity levels makes a \cross{reasonable} \fei{good} trade-off between accuracy and \cross{complexity} \fei{efficiency}. 

\subsection{Weight representations}


Due to the subset relationship between the small and tiny sub-models, storing both sub-models consumes almost the same storage as storing the small sub-model along.
The sparse matrices can be represented in many different formats, such as compressed sparse row (CSR), compressed sparse column (CSC), dictionary of keys (DOK), list of list (LIL), coordinate list (COO)~\cite{Langr2016SparseMatrix}. 
We use CSR~\cite{CSR1967IEEE} to describe the efficiency of storing both small and tiny sub-models, but the same approach can be easily adapted to other formats.


The CSR format represents a sparse matrix $A$ using three arrays, called $data$, $index$, and $ind\_ptr$. 
Assume $A$ has $m$ rows and $n$ columns, and the number of non-zero elements in $A$ is $N$, the array $data$ stores all the numerical non-zero values of $A$ in the row-scanning order and thus the length of $data$ is $N$. 
The $index$ array stores the column indices of the non-zero elements with the same order as in $data$, and thus the length of $index$ is also $N$. 
The $ind\_ptr$ array, whose length is $m+1$, stores the index pointer pointing to the first value in each row in $data$ and $index$.
Note that state-of-the-art pruning algorithms is capable to prune modern DNNs by $2\times$ to $10\times$ with negligible accuracy loss. Hence, $N$ is still much larger than $m$ or $n$ and thus the size of $ind\_ptr$ is negligible.

The CSR \fei{format} can be illustrated by the following example. Suppose the sparse matrix $A$ in Eq~(\ref{eq:AB1})  \cross{corresponds to} \fei{is a weight matrix in} a small sub-model, $B$ in Eq~(\ref{eq:AB2}) \cross{corresponds to a} \fei{is a subset of $A$, referring to the weight matrix in the corresponding} tiny sub-model.
\begin{align}
A = 
\begin{bmatrix}\label{eq:AB1}
0 & 1 & 0 & 0 & 0 & 0 & 0 & 0  \\
2 & 0 & 0 & 8 & 0 & 0 & 7 & 0 \\
0 & 0 & 3 & 0 & 0 & 5 & 0 & 0  \\
0 & 0 & 0 & 0 & 9 & 0 & 6 & 4 
\end{bmatrix}, \\
B = 
\begin{bmatrix}\label{eq:AB2}
0 & 1 & 0 & 0 & 0 & 0 & 0 & 0  \\
0 & 0 & 0 & 8 & 0 & 0 & 7 & 0 \\
0 & 0 & 3 & 0 & 0 & 0 & 0 & 0  \\
0 & 0 & 0 & 0 & 0 & 0 & 6 & 0 
\end{bmatrix}, 
\end{align}

To represent $A$ by zero-based numbering, 
\begin{align*}
    & data = [1, 2, 8, 7, 3, 5, 9, 6, 4], 
    index = [1, 0, 3, 6, 2, 5, 4, 6, 7], \\
    & ind\_ptr = [0, 1, 4, 6, 9].
\end{align*}
To decode the CSR format, $ind\_ptr$ first splits $data$ and $index$ into rows by checking the number of non-zero elements in each row ($ind\_ptr[i] - ind\_ptr[i-1]$, for $i=1,\ldots,m$). In the above example, the four rows have $(1,3,2,3)$ non-zero elements, thus $data$ and $index$ are split into $[[1],[2,8,7],[3,5],[9,6,4]]$ and $[[1], [0, 3, 6], [2, 5], [4, 6, 7]]$, respectively.
Note that the element order in the same row can be arbitrary in standard CSR.

In order to represent a sub-matrix $B$ in Eq~(\ref{eq:AB1}) where $B \subset A$, we first permute the values in $data$ such that it is still in row-scanning order but the values in $B$ appear before those values in $A$\textbackslash $B$ in each row. 
\textcolor{black}{We also allocate an additional array $end\_ind\_ptr$ of length $m$. The entry $i$ in $end\_ind\_ptr$ stores the index of the last non-zero value plus one in the $data$ and $index$ array for row $i$.}

Thus, we have a compact representation of $A$ and $B$ as 
\begin{align*}
    & data = [1, 8, 7, 2, 3, 5, 6, 9, 4], 
    index = [1, 3, 6, 0, 2, 5, 6, 4, 7] , \\
    & ind\_ptr = [0, 1, 4, 6, 9],
    end\_ind\_ptr(B) = [1, 3, 5, 7], \\
    & end\_ind\_ptr(A) = [1, 4, 6, 9].
\end{align*}

Note that $end\_ind\_ptr(A)$ is the same as $ind\_ptr$ by removing the first element which is a constant 0. Thus it does not require a separate array, and can be stored as $end\_ind\_ptr$ with an offset one.
The decoding of $A$ (or $B$) first splits $data$ according to $ind\_ptr$ and $end\_ind\_ptr(A)$ (or $end\_ind\_ptr(B)$). For the $i$-th row, the starting index in $data$ is $ind\_ptr[i]$ and the ending index is $end\_ind\_ptr\_A[i]$ (or $end\_ind\_ptr\_B[i]$), exclusively. For example, to decode $B$, $data$ and $index$ are split into $[data[0:1), data[1:3), data[4:5), data[6:7)]=[[1],[8,7],[3],[6]]$ and $[index[0:1), index[1:3), index[4:5), index[6:7)]]=[[1],[3,6],[2],[6]],$ respectively. Then the non-zero values of each row in $B$ can be identified.

Since the storage cost of $ind\_ptr$ and $end\_ind\_ptr$ are both negligible compared with $data$ and $index$, using the proposed method to store both $A$ and $B$ (both the small and tiny sub-models) has almost the same cost of storing $A$ (the small sub-model). 
Also note that by adding more $end\_ind\_ptr$ arrays and re-permuting $data$ and $index$, we can increase the number of sparse sub-models to an arbitrary number while the the storage cost of all sub-models is almost the same as the least sparse sub-model.

\subsection{Dynamic inference rerouting between small and tiny sub-models}
The tiny-dense-small method creates two sparse sub-models with different weight sparsity and computation complexity. However, selecting a preferred sparsity level before each inference might not be sufficient for some time-critical applications with hard real-time constraint. 
If the processor unexpectedly slows down after the start of inferencing the small sub-model, it is preferred to detect such situation in the first few layers and 
{\em dynamically} reroute to the tiny sub-model for the rest of the layers to reduce the latency. 
\textcolor{black}{
As being said, the rerouting can be initiated if the first part of the small sub-model suffers extra unexpected latency and hope to complete the inference in time without totally throwing away the computation already done in the small sub-model. }

We propose a dynamic inference rerouting such that we can inference the first few layers in the small sub-model and use the intermediate results (activation or feature maps) to reroute to the corresponding layers in the tiny sub-model. 
We constrain the rerouting to be from small to tiny instead of the opposite direction, since it is preferred to inference with the small sub-model in the first place to maintain accuracy. Figure~\ref{fig:dynamic} illustrates our proposed dynamic rerouting.
To enable this functionality during training, if a rerouting location at the $i$-th layer is desired, we fix the small sub-model (including all parameters such that all weights, biases, batch normalization related weights, biases, running means, and running variances) until the $i$-th layer and train the tiny sub-model from the $(i+1)$-st layer to the end, where only biases and batch normalization related weights can be changed, {\it i.e.}, the weights in convolution, MV multiplications and MM multiplications are still following the weight hierarchy from the small sub-model.
Experiments show that the propose dynamic rerouting not only enables to handle unexpected latency during the first few layers of inference, but also makes a good trade-off between the accuracy and the computation complexity (e.g., FLOPs in deployment).

\section{Experiments}
We have evaluated the proposed methodology on four different learning tasks: 2-D image recognition, 3-D object recognition, machine translation, and transductive learning. \fei{All experiments are performed on the PyTorch~\cite{paszke2019pytorch} framework}. \textcolor{black}{The results are shown in Table~\ref{tab:results}, and they have validated the effectiveness of our methodology.} 

\subsection{2-D image recognition}

We use a popular deep convolutional neural network, ResNet-50 model, on ImageNet dataset to demonstrate the results on hierarchically structured sparse deep neural networks. 

The first section in Table~\ref{tab:results} shows our results compared to the dense ResNet-50. The tiny-dense-small method generates a tiny sub-model and a small sub-model by keeping 5.43\% and 20.36\% of the weights, respectively. The resultant percentage of FLOPS are 8.20\% and 22.69\%, respectively. It is observed that the accuracy of the small sub-model is slightly better than the dense counterpart~\cite{he2016deep} and the tiny sub-model reduces 91.8\% FLOPs with only 2.6\% top-1 accuracy loss.  Our tiny sub-model has 5\% of weights of a dense counterpart, which achieves 1.4\% higher top-1 accuracy than Slimmable neural network \cite{yu2019slimmable} with 27.06\% of weights of a dense counterpart. The results on dynamic rerouting is shown in the next four rows. These four results correspond to four different switching locations, which are after the first, second, third, and fourth residual blocks~\cite{he2016deep}. It is observed that switching at a later point better preserves accuracy while keeping higher percentage of weights and FLOPs. 

Note that the percentage of weights are not precisely equal to the FLOPs due to three main reasons. First, the first  convolutional layer before the residual blocks is not pruned but its weights are shared by all sub-models. Second, the distributions of weights and FLOPs across ResNet-50 layers are very different. Third, the biases and batch normalization related parameters (0.4\% of a dense model) are not pruned as they barely contribute to the FLOPs.

\begin{table*}[t]
\small
    \centering
    \begin{tabular}{c|c|c|c|c|c}
    \toprule [0.2em]
        Base model & Sub-models & \multicolumn{2}{|c|}{Accuracy} & 
        \begin{tabular}{c}
          Weights \\percentage 
        \end{tabular} & 
        \begin{tabular}{c}
          FLOPs \\ percentage \\
        \end{tabular}
            \\ \hline
        \multirow{9}{5em}{ResNet-50}  & Dense~\cite{he2016deep} & 76.10\% & 92.90\% & 100\% & 100\%\\ 
          & Slimmable NNs \cite{yu2019slimmable} & 72.10\% &  - & 27.06\% & 26.83\%  \\ 
          & Slimmable NNs \cite{yu2019slimmable} & 65.00\% &  - & 7.84\% & 6.78\%  \\ 
          & Small sub-model & 76.45\% &  93.10\% & 20.36\% & 22.69\%  \\ 
          & Tiny sub-model  & 73.50\% & 91.72\% & 5.43\%  & 8.20\%         \\  
          & Small-tiny reroute 1 & 73.66\% & 91.71\% & 5.56\% & 10.84\% \\ 
          & Small-tiny reroute 2 & 73.84\% & 91.94\% & 6.26\% & 14.48\%  \\  
          & Small-tiny reroute 3 & 74.41\% & 92.16\% & 10.40\% &19.99\% \\  
          & Small-tiny reroute 4 & 76.14\% &  93.04\% & 19.16\% & 22.69\% \\  \hline

        \multirow{4}{5em}{GNMT} & Dense~\cite{wu2016gnmt}  & \multicolumn{2}{|c|}{24.61} & 100\% & 100\%\\ 
          & Small sub-model  &  \multicolumn{2}{|c|}{24.47}  & 36.53\% & 20.02\% \\ 
          & Tiny sub-model  &  \multicolumn{2}{|c|}{23.06} &  24.63\% & 5.02\%\\ 
          & Small-tiny reroute  & \multicolumn{2}{|c|}{23.65} & 28.94\% & 10.45\%\\ \hline

        \multirow{3}{5em}{PointNet++} & Dense~\cite{qi2017pointnet++} & \multicolumn{2}{|c|}{91.9\%}  & 100\% & 100\% \\ 
          & Small sub-model  & \multicolumn{2}{|c|}{91.94\% $\pm$ 0.14\%}  & 4.25\% &  3.00\% \\ 
          & Tiny sub-model  &   \multicolumn{2}{|c|}{89.92\% $\pm$ 0.21\%} & 1.99\%  &  0.72\%    \\ \hline

        \multirow{3}{5em}{GAT} & Dense ~\cite{velickovic2018gat} & \multicolumn{2}{|c|}{83.00\%} & 100\%  & 100\%\\ 
          & Small sub-model  & \multicolumn{2}{|c|}{84.80\%}  & 10.14\% & 36.81\% \\
          & Tiny sub-model  &   \multicolumn{2}{|c|}{81.00\%} & 3.15\%  & 13.48\%       \\ 
    \bottomrule [0.2em]
    \end{tabular}
    \caption{\small Accuracy of dense, small, and tiny sub-models of ResNet-50, GNMT, PointNet++, and GAT. ResNet-50s show top-1 and top-5 accuracy, GNMTs show BLEU scores, PointNet++'s and GATs show top-1 accuracy.}
    \label{tab:results}
\end{table*}

\subsection{Machine translation}
We use Google neural machine translation (GNMT)~\cite{wu2016gnmt} model on WMT En $\rightarrow$ De dataset to demonstrate the proposed method on  machine translation tasks.
GNMT consists of eight layers of long short-term memory (LSTM) in both encoder and decoder and a classifier. GNMT contains over 127 millions of parameters excluding 33 millions word embedding, and the proportion of MV and MM multiplication related weights is 99.9\%. The word embedding is not pruned because pruning it may not easily translate to FLOP savings. In our experiments, the word embedding is shared between the encoder and the decoder, as well as the small and the tiny sub-models.

We prune 80\% of the prunable weights for the small sub-model, resulting to 36.53\% of the weights. The main difference is due to the word embedding, which remains dense. The FLOPs of the small sub-model is reduced to 20.02\% of the dense model. The 0.02\% extra FLOPs is due to the small attention matrix multiplications, which remains dense. The small sub-model slightly decreases the BLEU score from the dense model (from 24.61 to 24.47).

Similarly, we prune 95\% of the prunable weights for the tiny sub-model, resulting to 24.63\% of the weights and 5.02\% of the FLOPs. It reduces the BLEU score to 23.06. When the model is rerouted from the small sub-model in the encoder to the tiny sub-model in the decoder, the remaining weights are 28.94\% and the remaining FLOPs are 10.45\%. The FLOPs  is calculated assuming the input and output sequence length of 16 with a beam size of one. 
Its BLEU score (23.65) falls between the tiny and small sub-models.

\subsection{3-D object recognition}
We use PointNet++ with multi-scale grouping (MSG)~\cite{qi2017pointnet++} on ModelNet40~\cite{wu2015shapenet} dataset for the 3-D object recognition. 
PointNet++ with MSG consists of several $1\times1$ convolutional layers (MM multiplications) and three fully connected layers (MV multiplications). 

Compared to the dense PointNet++~\cite{qi2017pointnet++}, the tiny-dense-small method generates a tiny sub-model and a small sub-model by keeping 1.99\% and 4.25\% of weights, respectively. The resultant percentage of FLOPS is reduced to 0.72\% and 3\% of their dense counterparts, respectively. Even larger difference is observed in the distribution of weights and FLOPs in PointNet++ layers. Since PointNet++ MSGs involves random sampling operations, we test the model 50 times and report their means and standard deviations. The resultant small sub-model is again slightly better than its dense counterpart.


\subsection{Transductive learning on graphs}
We use graph attention networks (GATs)~\cite{velickovic2018gat}  on the Cora citation dataset~\cite{Sen2008Cora} to demonstrate the propose method on transductive learning tasks. 
The GATs 
consist of eight attention heads and an output classifier, all of which are MV and MM multiplications. 
We follow the same setup~\cite{velickovic2018gat} in partitioning training nodes, validation nodes, and test nodes. 

We observe from Table~\ref{tab:results} (last three rows) that the proposed tiny-dense-small method generates a small sub-model with roughly 10\% of weights and 36\% FLOPs while the accuracy is improved compared to the dense counterpart. A more-sparse tiny GAT sub-model has about 3\% of weights and 13\% FLOPs and the accuracy is only reduced by 2\%. The difference in percentage of weights and FLOPs is due to the attention layers which remain dense.

\subsection{Weight savings compared with two separate sparse models}
\begin{figure} [t]
  \centering
  \includegraphics[width=1\linewidth]{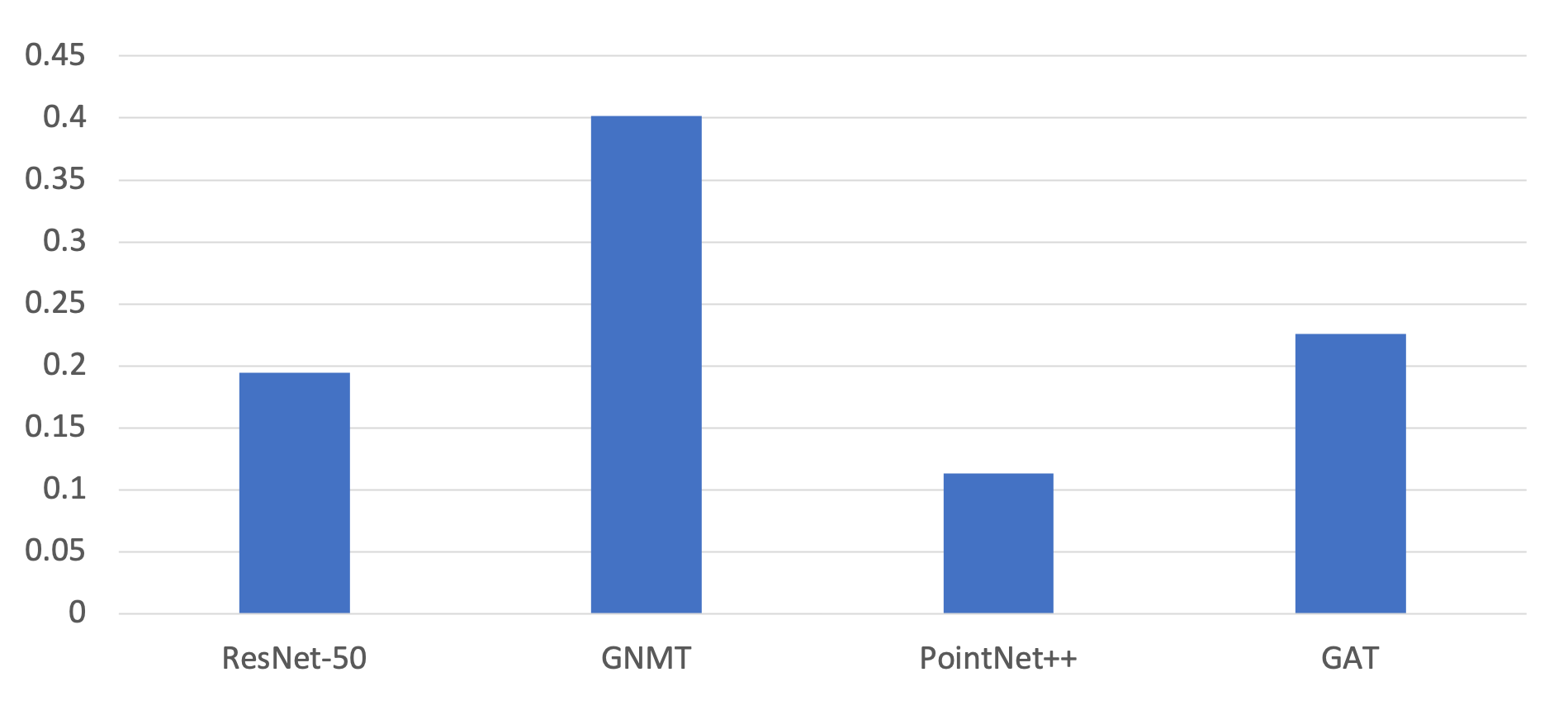}
  \caption{{\small \textcolor{black}{Percentage of weights savings of the sub-models with weight hierarchy over two separate sparse models (small and tiny).}}}
  \label{fig:wt_savings} 
\end{figure}

\textcolor{black}{
A benefit of introducing weight hierarchy is to reduce the memory and storage cost of storing multiple sub-models. If there is no weight hierarchy, we need to store small and tiny models separately.
In Figure~\ref{fig:wt_savings}, we present the percentage of weight savings when using hierarchical sub-models
over using separate models. 
An average of 20\% to 30\% of storage reductions are expected from saving two models separately. The overall storage savings may be larger if CSR format is used and the storage for the $index$ and $ind\_ptr$ arrays are counted. It should also be noted that the savings are amplified when more than two sparsity levels are desired. }

\begin{figure} [t]
  \centering
  \includegraphics[width=1\linewidth]{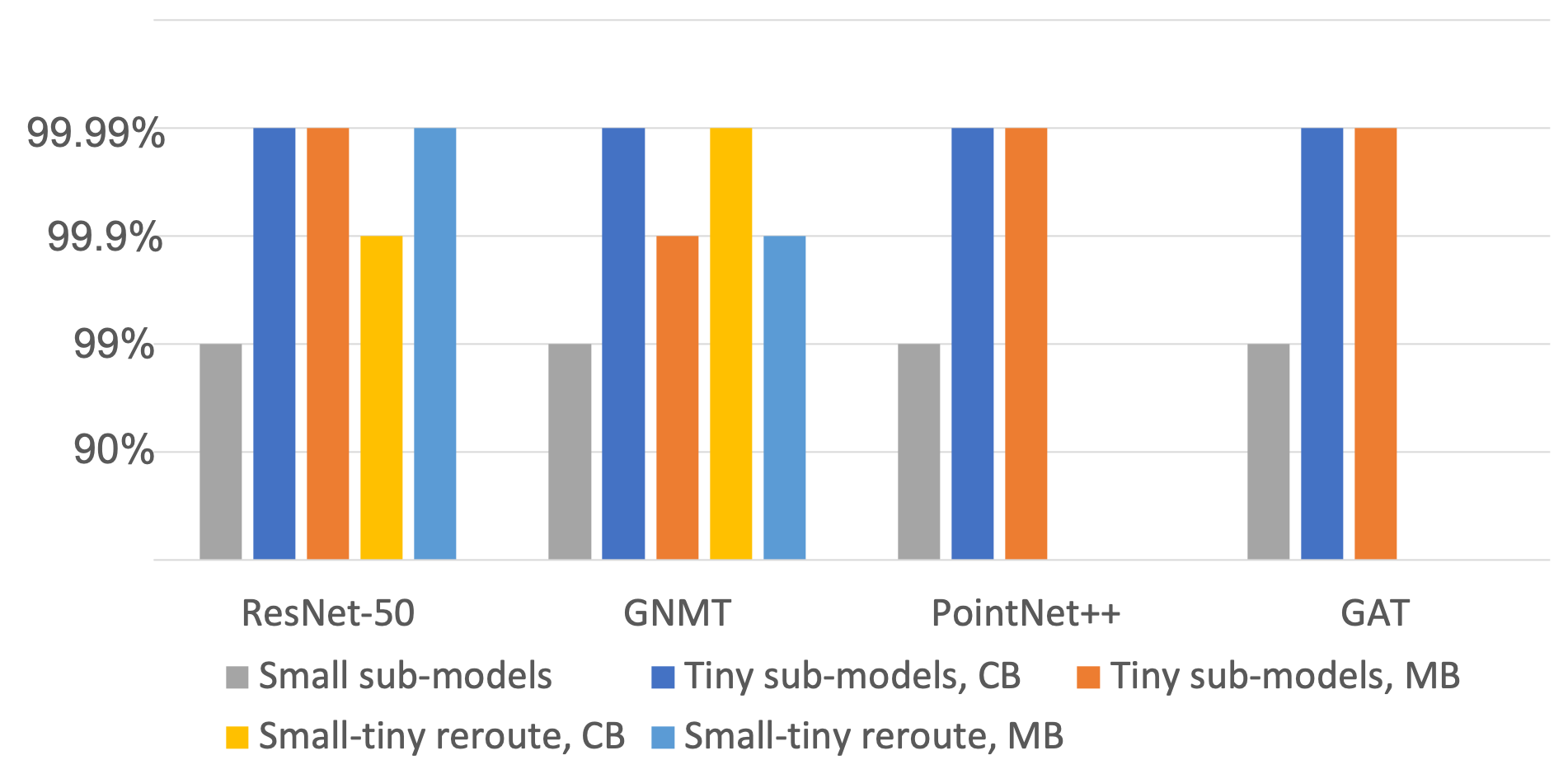}
  \caption{{\small{QoS comparison of small sub-models, tiny sub-models, and small-to-tiny reroutes  in computation-bound (CB) or memory-bound (MB) scenarios.}}}
  \label{fig:qos} 
\end{figure}

\subsection{Quality of Service (QoS) improvement}
In order to verify that dynamically selecting the small and tiny sub-models before and during inferences can improve the quality of service, we have collected the the latency of over one million inferences on iPhone X devices on a proprietary model~\cite{alibaba2020mnn}. The distribution of the latency is similar to the one reported in~\cite{wu2019machine}. 
We hypothesize that the latency distributions of the models reported in Table~\ref{tab:results} are similar to the collected distribution. We further linearly correlate the number of FLOPs and the number of weight parameters to the execution latency, depending on whether the inference is computation-bound (CB) or memory-bound (MB). 
We estimate that for the inference latency corresponding to the 99\textsuperscript{th} percentile of the small sub-models, the tiny sub-models are capable to reach 99.99\textsuperscript{th} percentile~\textcolor{black}{most of the time, on either computation-bound or memory-bound inference systems}, as shown in Figure~\ref{fig:qos}. \mh{The quality-of-service (QoS) reflects the percentage of inferences (e.g., 99.99\%) within a certain latency requirement. }
Our result illustrates that if the decision to inference the small or tiny sub-model is made before each inference based on the runtime condition, the QoS can be improved by two nines {(\it i.e.}, from 99\% to 99.99\%). Similarly, in the case that the runtime condition is changed during inference and the inference is rerouted from small sub-models to tiny sub-models, the QoS can still be improved by at least one nine {(\it i.e.}, from 99\% to 99.9\%). 
This experiment on QoS estimation has preliminarily validated the usefulness of our approach in improving the QoS in the wild.


\subsection{Comparison to existing dynamic DNNs}
Several recent works \cite{wang2018skipnet,yu2019slimmable,yang2019condconv,ma2020weightnet} focus on \cross{``conditional''} \fei{``input dependent''} dynamic inference, i.e., \cross{which} \fei{the} path to inference is determined by \fei{the} individual \cross{input samples} \fei{input, } where \fei{the} ``easy'' inputs go through simple paths and thus reduce computation\cross{s} on average. \cross{A fundamental difference of} Our ``dynamic rerouting'' is \fei{fundamentally different in} that ours is {input independent}, i.e., {any} input \cross{sample} \fei{data} can be switched from small to tiny sub-model\fei{s} from {any} designated switching point\cross{s} (e.g., 4 points for ResNet50). The switch\fei{ing} decision is based on \fei{the} run-time status such as \cross{current} \fei{the} latency before each switch\fei{ing} point, independent of \cross{samples} \fei{the input data}. Input independent rerouting is much simpler and more practical than \fei{the} NN based conditional rerouting in existing literature\fei{s}.
Second, our DNN models are much more sparse than the existing works. For example, our tiny ResNet50 has 5\% of \fei{the} weights of \cross{a} \fei{the} dense counterpart, while SkipNet \cite{wang2018skipnet} has 88\%. Slimmable neural network \cite{yu2019slimmable} has one result \cross{on 27.06\% weights (and 1.4\% less accurate than our tiny ResNet50)} \fei{that achieves 1.4\% {\em less} accuracy than our tiny ResNet50 with 5$\times$ parameters (27.06\% vs 5.43\%)}. CondConv \cite{yang2019condconv} or WeightNet \cite{ma2020weightnet} \cross{is} \fei{are} even larger than the dense ResNet50.

\section{Conclusions}
Realized by the large fluctuations in the runtime environment and the difficulty of maintaining QoS with one model due to the long tail inferences with large latency, we propose a compact multi-level sparse neural network with a hierarchical structure of weights such that the more-sparse sub-model is a subset of the less-sparse sub-model. This hierarchy enables us to select the sub-model of the desired sparsity level before and during each inference, while the storage cost of multiple sub-models is almost the same as the least sparse one. 
Experiments show that the small sub-models can reduce the number of weights and FLOPs by an average of 7$\times$ without accuracy degradation and a further 3$\times$ reduction can be realized at the cost of only 2-3\% of relative accuracy. 
In this work, we have validated our methodology from the algorithm perspective. Our next step is to translate the  gain into latency reduction via structured pruning and hardware/software co-design of the inference system.


\bibliographystyle{IEEEtran}
\bibliography{reference}


\end{document}